\documentclass[10pt,twocolumn,letterpaper]{article}

\usepackage{cvpr}             
\usepackage[table]{xcolor}
\usepackage{bigstrut}
\usepackage{utfsym}
\usepackage{tcolorbox}
\usepackage{multirow}
\usepackage{bm}
\usepackage{graphicx}

\definecolor{cvprblue}{rgb}{0.21,0.49,0.74}
\usepackage[pagebackref,breaklinks,colorlinks,citecolor=cvprblue]{hyperref}

\title{SR-CIS: Self-Reflective Incremental System with Decoupled Memory and Reasoning}
\vspace{-10pt}

\author{
  Biqing Qi\textsuperscript{1,4}, 
  Junqi Gao\textsuperscript{2,}\thanks{Corresponding authors: Bowen Zhou and Junqi Gao.},
  Xinquan Chen\textsuperscript{2},
  Dong Li\textsuperscript{2},
  Weinan Zhang\textsuperscript{3},
  Bowen Zhou\textsuperscript{1,4,}\footnotemark[1] \\
  $^1$ Department of Electronic Engineering, Tsinghua University, \\
  $^2$ School of Mathematics, Harbin Institute of Technology, \\
  $^3$ Faculty of Computing, Harbin Institute of Technology, \\
  $^4$ Shanghai AI Laboratory, \\
  {\tt\small \{qibiqing7,gjunqi97,xinquanchen0117,arvinlee826\}@gmail.com,} \\ 
  {\tt\small \{wnzhang\}@ir.hit.edu.cn, \{zhoubowen\}@tsinghua.edu.cn}
  }

\begin{document}
\newtcolorbox{mycolortbox}{
    colback=blue!5, 
    colframe=black, 
    arc=4pt, 
    boxrule=1pt, 
    fonttitle=\bfseries
}
\newtcolorbox{mytexttbox}{
    colback=yellow!5, 
    colframe=black, 
    arc=4pt, 
    boxrule=1pt, 
    fonttitle=\bfseries
}
\maketitle
\begin{abstract}
\vspace{-5pt}
  The ability of humans to rapidly learn new knowledge while retaining old memories poses a significant challenge for current deep learning models. To handle this challenge, we draw inspiration from human memory and learning mechanisms and propose the Self-Reflective Complementary Incremental System (SR-CIS). Comprising the deconstructed Complementary Inference Module (CIM) and Complementary Memory Module (CMM), SR-CIS features a small model for fast inference and a large model for slow deliberation in CIM, enabled by the Confidence-Aware Online Anomaly Detection (CA-OAD) mechanism for efficient collaboration. CMM consists of task-specific Short-Term Memory (STM) region and a universal Long-Term Memory (LTM) region. By setting task-specific Low-Rank Adaptive (LoRA) and corresponding prototype weights and biases, it instantiates external storage for parameter and representation memory, thus deconstructing the memory module from the inference module. By storing textual descriptions of images during training and combining them with the Scenario Replay Module (SRM) post-training for memory combination, along with periodic short-to-long-term memory restructuring, SR-CIS achieves stable incremental memory with limited storage requirements. Balancing model plasticity and memory stability under constraints of limited storage and low data resources, SR-CIS surpasses existing competitive baselines on multiple standard and few-shot incremental learning benchmarks.
  \vspace{-5pt}
\end{abstract}    
\section{Introduction}
\label{sec:intro}

\begin{table*}[t]
\setlength{\tabcolsep}{1pt} 
  \renewcommand{\arraystretch}{0.25}
    \centering
    \small
    \caption{Advantages and disadvantages of existing CL strategies}
    \begin{tabular}{cccccc}
        \toprule[1pt]
      \textbf{Type} & \textbf{Strategy} & \textbf{Storage Limitation} & \textbf{Low Data Resource} & \textbf{Memory Stability}  & \textbf{Model Plasticity}\\
        \midrule[0.5pt]

        \multirow{2}[1]{*}{Rehearsal} & Real-data rehearsal & \textcolor{red}{\usym{2718}} & \textcolor{blue}{\usym{2714}} & \textcolor{blue}{\usym{2714}} & \textcolor{blue}{\usym{2714}} \\
        \cmidrule[0.5pt]{2-6}
         & Pseudo rehearsal & \textcolor{blue}{\usym{2714}} & \textcolor{red}{\usym{2718}} & \textcolor{blue}{\usym{2714}} & \textcolor{blue}{\usym{2714}}\\
        \midrule[0.5pt]
        \multirow{2}[1]{*}{Parameter allocation}
        & Static Parameter Assignment & \textcolor{blue}{\usym{2714}} & \textcolor{blue}{\usym{2714}} & \textcolor{blue}{\usym{2714}} & \textcolor{red}{\usym{2718}}\\
         \cmidrule[0.5pt]{2-6}
        & Dynamic Parameter Expansion & \textcolor{red}{\usym{2718}} & \textcolor{blue}{\usym{2714}} & \textcolor{blue}{\usym{2714}} & \textcolor{blue}{\usym{2714}}\\
         \midrule[0.5pt]
         Regularization
        & Parameter Regularization & \textcolor{blue}{\usym{2714}} & \textcolor{blue}{\usym{2714}} & \textcolor{blue}{\usym{2714}} & \textcolor{red}{\usym{2718}}\\
        \bottomrule[1pt]
    \end{tabular}
    \label{tab:trade-off}
    % \vspace{-5pt}
\end{table*}

Deep learning models have demonstrated superior performance in a wide range of downstream tasks \cite{HeZRS16,HoJA20,LiuLWL23a}. However, when faced with tasks involving the continuous input of new classes, these models often rapidly forget old task knowledge during the learning process on new tasks, a phenomenon known as catastrophic forgetting \cite{French93, KemkerMAHK18}. In contrast, humans can continuously learn new knowledge while retaining old knowledge.

\noindent\textbf{Related Works and Considerations}
To bridge this gap, Continual Learning (CL) has been proposed to develop deep learning models that can continuously learn from new data without forgetting previous knowledge, thus addressing catastrophic forgetting, which is crucial for adapting to changing task scenarios in real-world applications \cite{RebuffiKSL17,YanX021}. Mainstream CL strategies can be categorized into three types: 1) maintaining a memory buffer to store data from current tasks and replaying them in subsequent tasks \cite{RolnickASLW19,AljundiLGB19}, or directly training incremental generative models to generate data for pseudo-replay \cite{CongZLWC20,GaoL23a}; 2) imposing regularization constraints on weight changes during updates \cite{KirkpatrickPRVD16,Adel0T20}; and 3) dedicating isolated parameters for different tasks to mitigate the impact on parameters from previous tasks during updates, which can be achieved by dynamically expanding additional parameters for new tasks \cite{RusuRDSKKPH16,AljundiCT17} or by allocating static fixed parameters to different tasks \cite{MallyaL18,AhnCLM19}.
However, static parameter allocation and the addition of regularization constraints often constrain the learning of new tasks implicitly or explicitly \cite{ParisiKPKW19}. Additionally, storing data for each class in a memory buffer or continuously expanding parameters for new tasks can result in significant storage overheads when dealing with continual task streams \cite{abs-2302-03648}. Although methods employing generative models for pseudo-replay are available, they come with certain limitations. These methods either require fine-tuning conditional generative models for each class in each task, which demands a large amount of training data to be effective \cite{abs-2302-00487}, or they need well-trained pre-trained classifiers on previous classes to guide the generation process \cite{GaoL23a}, which is often impractical.
In this work, we aim to integrate the advantages of existing CL strategies while avoiding their limitations. Our goal is to balance stable memory for old tasks and plasticity for new tasks within fixed storage and low data resource constraints. Drawing inspiration from human memory and learning, we consider the brain's organization, where multiple regions learn and remember complementarily \cite{Zilles2010CentenaryOB}. The Complementary Learning Systems (CLS) theory provides a clear framework for this insight \cite{McClelland1995WhyTA}. According to this theory, the brain employs two complementary systems, the hippocampus and the neocortex, to cooperatively handle learning and memory. The hippocampus is responsible for storing and encoding short-term, context-specific memories, while the neocortex integrates and stores structured long-term memories.

\begin{figure*}[h]
    \centering
    \includegraphics[width=0.8\textwidth]{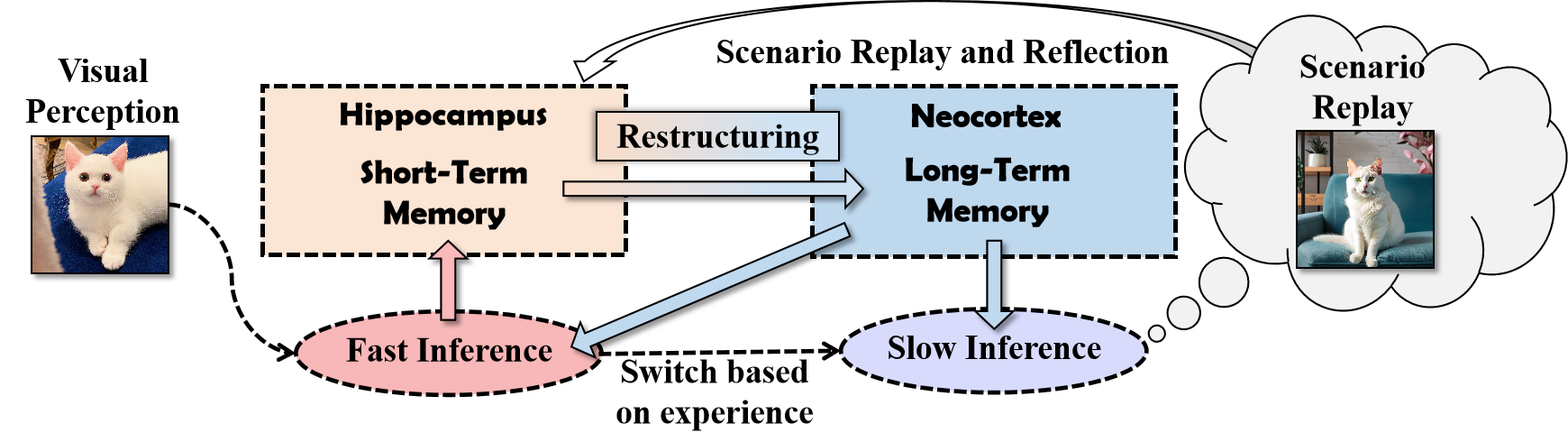}
    \caption{The schematic diagram of an ideal complementary learning system.}
    \label{Illustration}
\end{figure*}

Currently, some CL works attempt to design strategies based on the CLS. However, most of these strategies are limited to either specifying two complementary tasks and global components for the CL model \cite{0002ZESZLRSPDP22} or simply instantiating two models for fast-slow updates \cite{AraniSZ22}. The work \cite{abs-2403-02628} first attempted to instantiate a fast-slow reasoning system through the collaboration of large and small models, but their design still has limitations:
1) Their system only emphasizes fast-slow reasoning at the inference level. In contrast, the human brain has specialized regions for memory and cognition working in synergy \cite{Luppi2022ASC}, suggesting that fast-slow memory and fast-slow reasoning should be deconstructed and complementary.
2) They only store class prototypes as memory, while the task transformation layer following the pre-trained backbone continues to update. This leads to the transformation layer still requiring real data rehearsal to maintain stable class prototype alignment.
3) They completely freeze the pre-trained backbone, severely limiting the model's plasticity, making it difficult to adjust when real data deviates significantly from the pre-training dataset.
4) Their fast-slow reasoning switch is only executed based on hard sample detection on a single batch, introducing uncertainty and affecting the accuracy of hard sample detection, leading to excessive unnecessary switches and inference burdens.
Given these considerations, we propose that the complementary system we aim to build should have the following three features, as illustrated in Fig. \ref{Illustration}:
\begin{itemize}
    \item \textbf{Decomposed short-long memory and fast-slow reasoning components}. This approach explicitly stores memories, enabling rapid learning of short-term task memories while regularly integrating these short-term memories of different tasks into universal long-term memories.
\item \textbf{Capability of restructuring short-term to long-term memory through regular scenario replay and reflection}. Similar to the collaboration between the hippocampus and neocortex, regular scenario replay and reflection are crucial for restructuring fragmented short-term memories into complete long-term memories \cite{OReilly2014ComplementaryLS}. This replay often does not require the reproduction of real data but involves scenario reenactment within memory: humans do not need to repeatedly view a person's photo to memorize their appearance.
\item \textbf{Ability to accurately judge when to switch to slow reasoning based on prior experience}. After accumulating task experience, humans can accurately assess task difficulty levels, relying on intuitive fast thinking for simple tasks and deep deliberation for more challenging ones. Fast and slow reasoning work together organically, efficiently completing a series of tasks \cite{Kramer2011}.
\end{itemize}

\noindent\textbf{System Design}
To construct such a system, we first formalize the system framework to clarify the specific functionalities required for each component. This step facilitates concrete component instantiation and method design. Based on this framework, we propose the Self-Reflective Complementary Incremental System (SR-CIS), which consists of a deconstructed Complementary Inference Module (CIM) and Complementary Memory Module (CMM). The CIM includes a Large Multimodal Model (MLLM) for executing slow reasoning, instantiated by LLaVA \cite{LiuLWL23a}, and a small model for fast inference, instantiated by pre-trained Vision Transformer (ViT) \cite{DosovitskiyB0WZ21}. To execute accurate mode switches in reasoning, we introduce the Confidence-Aware Online Anomaly Detection (CA-OAD) mechanism. This mechanism merges each sample's confidence and its standard deviation online through Exponential Moving Average (EMA) and performs hard sample assessment. During training, the temperature scaling factor is tuned through feedback to penalize irrational behaviors, such as screening out high-confidence negative samples and low-confidence positive samples, ensuring accurate anomaly detection.
CMM consists of a Short-Term Memory (STM) region corresponding to the current task and a Long-Term Memory (LTM) region obtained from integrating task memories. Each memory region is divided into parameter memory instantiated by Low-Rank Adaptation (LoRA) \cite{HuSWALWWC22} corresponding to the task and representation memory instantiated by prototype weights and biases. To facilitate periodic STM restructuring, we establish a scenario description pool for efficient storage of text scenario descriptions, along with a Scenario Replay Module (SRM) instantiated by Stable-Diffusion (SD) for memory replay.

For each task, we fine-tune the small model with task-specific LoRA to ensure its plasticity. We allocate independent prototype weights and memory biases for each class. To achieve this, we design a prototype alignment loss with temperature scaling, updating relevant prototype weights and memory biases to form task-specific STM. Meanwhile, the MLLM in CIM describes a small amount of training data and adds it to the scenario description pool.

During the inference stage, we first conduct memory restructuring. By uniformly sampling language descriptions from the description pool as prompts, we use SD for scenario replay of previous task samples. This combines LoRA of different tasks, forming task-agnostic parameter memory. Because scenario replay is limited to memory consolidation, it does not introduce parameter bias towards specific tasks. To ensure storage boundedness, we periodically merge the combined parameter memory into the LTM region and delete the corresponding short-term parameter memory. Additionally, CA-OAD performs online monitoring of predictions from the small model, feeding back selected hard samples to the large model for slow inference. This effectively integrates fast and slow reasoning.

SR-CIS is the first to propose a complete memory-inference deconstructed complementary learning system. It combines the advantages of various CL strategies, balancing model plasticity and memory stability with limited storage requirements and data resources. In standard and few-shot class incremental learning experiments on multiple benchmarks, our approach outperforms a range of current competitive baselines.
\section{Problem Formulation}
A standard class-incremental learning problem consists of a series of disjoint tasks $\mathcal T_{1},\mathcal T_{2},\dots, \mathcal T_{T}$. Each task $\mathcal T_{t}$ comprises input sample pairs $\bm z^{(t)} = (\bm x^{(t)}, y^{(t)})$ which are i.i.d. samples drawn from distribution $\mathcal D_t$, forming the corresponding training set $\mathcal S_{t}^{tr}=\{(\bm x_i^{(t)}, y_i^{(t)})\}_{i=1}^{|\mathcal S_{t}^{tr}|}$ and test set $\mathcal S_{t}^{te}=\{(\bm x_i^{(t)}, y_i^{(t)})\}_{i=1}^{|\mathcal S_{t}^{te}|}$. Here, $y^t\in\mathcal Y^t$ denotes the label set corresponding to $\mathcal T_t$, with $\mathcal Y^{t_1}\cap\mathcal Y^{t_2}=\emptyset, \forall t_1\ne t_2$. During the incremental training phase, the incremental training set $\mathcal{S}{t}$ is sequentially inputted for model training. Upon completion of training on $\mathcal{S}_{t}^{tr}$, all the seen training sets $\mathcal{S}_{1}^{tr}, \dots, \mathcal{S}_{t}^{tr}$ are no longer visible. The incremental model is evaluated on the union of test sets from all previously encountered tasks, denoted as $\mathcal E_{t} = \cup_{i=1}^{t}\mathcal S_{i}^{te}$.
\section{SR-CIS: A CLS with Memory and Reasoning Deconstruction}
In this section, we first formalize the system components based on their expected functionalities and introduce the three main actions of the system. Next, we detail the execution of these actions while presenting the instantiation of each component.
\subsection{System Formulation}
\label{sec 2.1}
As mentioned earlier, an ideal CLS should consist of two deconstructed parts, namely the CMM $\mathcal M$ and CIM $\mathcal I$. The parameter memory $\mathcal M_p$ and representation memory $\mathcal M_r$ learned in tasks form the LTM region $\mathcal M^{LTM} = \{\mathcal M_p^{LTM}, \mathcal M_r^{LTM}\}$ and STM region $\mathcal M^{STM} = \{\mathcal M_p^{STM}, \mathcal M_r^{STM}\}$. Together with the scenario memory $\mathcal M^{Sce}$ used for recall and memory restructuring, they constitute the CMM $\mathcal M = \{\mathcal M^{LTM}, \mathcal M^{STM}, \mathcal M^{Sce}\}$. The CIM consists of a fast inference component $\mathcal I^{F}$, a slow inference component $\mathcal I^{S}$, and an inference mode switching mechanism $\mathcal I^{Swi}$, represented as $\mathcal I = \{\mathcal I^{F}, \mathcal I^{S}, \mathcal I^{Swi}\}$. With these components in place, the system can perform the following actions:

\noindent\textbf{Action $\mathcal A_{L}$: Learning.}
$\mathcal I^{F}$ receives the training set $\mathcal{S}_{t}^{tr}$ and performs training. During the training process, a part of samples $\bm z^{(t)}$ are memorized as scenarios $\bm s^{(t)}$ through the scenario recording operation $\mathcal R_{\mathcal M}$, and added to the scenario memory: $\mathcal M^{Sce} = \mathcal M^{Sce}\cup \bm s^{(t)}$, where $\bm s^{(t)} = \mathcal R_{\mathcal M}(\bm x^{(t)}, y^{(t)})$. Additionally, each training sample to the accumulation of online classification experience, denoted as $\mathcal W$, which $\mathcal I^{Swi}$ references during testing. Upon completion of training, the resulting short-term parameter memory $\mathcal M_p^{STM,t}$ and representation memory $\mathcal M_r^{STM,t}$ corresponding to the task are stored in STM: $\mathcal M_p^{STM} = \mathcal M_p^{STM}\cup\mathcal M_p^{STM,t}$, $\mathcal M_r^{STM} = \mathcal M_r^{STM}\cup\mathcal M_r^{STM,t}$. $\mathcal M_p^{STM}$ and $\mathcal M_r^{STM}$ are initialized as $\emptyset$.

\noindent\textbf{Action $\mathcal A_{R}$: Memory Restructuring.}
Using the scenarios in $\mathcal M^{Sce}$ for scenario replay, we obtain a set of scenarios $\tilde{\mathcal S}=\{(\tilde{\bm x}_i^{(t)}, y_i^{(t)})\}_{i=1}^{N}$, where the replayed scenario $\tilde{\bm x}_i$ is obtained through the scenario reproduction operation $\tilde{\mathcal R}_{\mathcal M}$: $\tilde{\bm x}_i = \tilde{\mathcal R}_{\mathcal M}(\bm s_i^{(t)})$. Subsequently, the restructuring mechanism $\mathcal U$ conducts memory restructuring: $\mathcal M_\Sigma^{STM}=\mathcal U(\mathcal M_r^{STM},\mathcal M_p^{STM},\tilde{\mathcal S})$, and then merged as LTM: $\mathcal M_r^{LTM}=\mathcal M_{\Sigma,r}^{STM}, \mathcal M_p^{LTM}=\mathcal M_{\Sigma,p}^{STM}$. When the number of elements in $\mathcal M_p^{STM}$, denoted as $\left|\mathcal M_p^{STM}\right|$, equals the restructuring period $e$, reset is performed: $\mathcal M_p^{STM} = \mathcal M_p^{LTM}$.

\noindent\textbf{Action $\mathcal A_{I}$: Fast and Slow Inference.}
During the inference phase, for each test input $\bm{x}^{(t)}$ with an unknown label, the fast inference is first executed as $o^{(t)} = \mathcal{I}^{F}(\bm{x}^{(t)})$. Then, the inference switching mechanism $\mathcal{I}^{Swi}$, performs anomaly detection combined with previous classification experience $\mathcal{W}$: $o^{Swi} = \mathcal{I}^{Swi}(o^{(t)}, \mathcal{W})$. If $o^{Swi} = 0$, the prediction result from $\mathcal{I}^{F}$ is directly returned as $\hat{y}$. Otherwise, the slow inference is conducted: $\hat{\hat{y}} = \mathcal{I}^{S}(o^{(t)})$, and the result is returned.

\subsection{STM Learning with Scenario Memorization}

\textbf{Instantiation of Fast and Slow Inference Components} 
In common understanding, fast thinking is more intuitive, while slow thinking is more rational. Based on this characteristic, we instantiate the fast inference component $ \mathcal{I}^{F}_{\bm{\theta}} $ using a classification model parameterized by $ \bm{\theta} $ for classification tasks. For the slow inference component $ \mathcal{I}^{S} $, we use an MLLM to perform more complex "rational" judgments. Specifically, deploying pre-trained models and fine-tuning them for downstream tasks is highly effective in current practices \cite{Chen000DLMX0021, abs-2003-08271}. The use of a pre-trained ViT architecture for incremental learning is also becoming mainstream \cite{0002ZESZLRSPDP22, abs-2404-00228}. Thus, we choose a pre-trained ViT as $\mathcal{I}^{F}_{\bm{\theta}}$. For $\mathcal{I}^{S}$, we select LLaVA, the most mainstream MLLM architecture \cite{LiuLWL23a}.

\noindent\textbf{Instantiation of Parameter and Representation Memory} 
After instantiating the inference components, we aim to 1) achieve memory deconstruction within the system and 2) ensure the model's plasticity for new tasks. LoRA enables efficient fine-tuning of pre-trained models, and its parameters can be explicitly stored and accessed. Therefore, we instantiate the task-specific parameter memory using LoRA, represented as $\mathbf{W}^{t} = \mathbf{B}^{(t)} \mathbf{A}^{(t)}$, to be merged into $\mathcal{I}^{F}$ during training. For each newly encountered class $y \in \mathcal{Y}^{uns}$, where $\mathcal{Y}^{uns}$ denotes the set of unseen classes, we allocate a set of prototype weights $\bm{p}_w^y$ and prototype biases $\bm{p}_b^y$ instantiated by learnable embeddings as representation memory.
\begin{figure*}[t]
    \centering
    \vspace{-10pt}
    \includegraphics[width=1.0\textwidth]{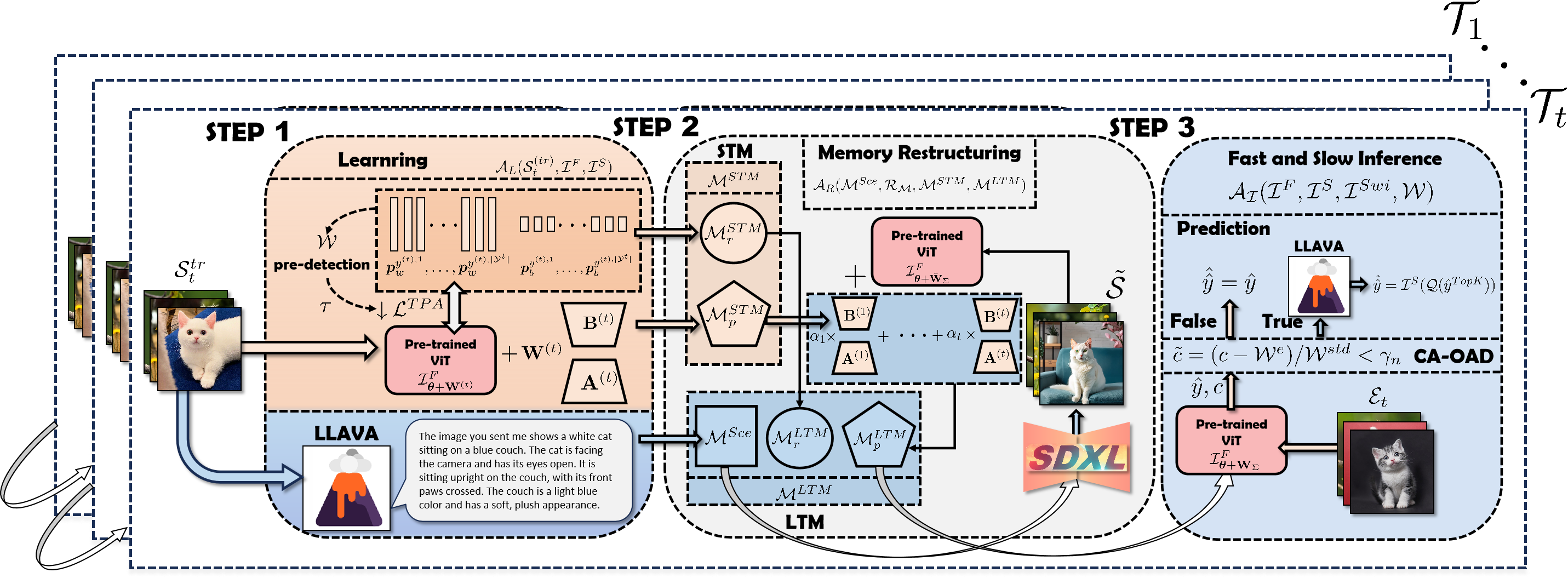}
    \vspace{-10pt}
    \caption{The schematic diagram of an ideal complementary learning system.}
    \label{Workflow}
\end{figure*}

\noindent\textbf{Scenario Recording of SRM}
To efficiently record scenarios corresponding to a portion of input samples for subsequent memory restructuring, we draw inspiration from multisensory learning \cite{Okray2023MultisensoryLearning}. For humans, directly recalling a specific scenario may be challenging, but it is easier to recall based on a description of that scenario. Motivated by this, we use $\mathcal{I}^{S}$ to generate textual descriptions $\bm{s} = \mathcal{R}_{\mathcal{M}}(\bm{x}, y)$ for $m$ samples of each class $(\bm{x}, y)$. These descriptions are then added to the scenario description pool: $\mathcal{M}^{Sce} = \mathcal{M}^{Sce} \cup \bm{s}$. This approach avoids the direct storage of raw images and instead stores scenarios in a more efficient textual form, significantly reducing storage consumption.

\noindent\textbf{Experience Accumulation for Anomaly Detection} 
On task $\mathcal T_t$, let the Softmax confidence denoted as $ \bm o^{(t)}$ and $c^{(t)}=\max_{1\le i\le |\mathcal Y^{see}|}\bm o^{(t)}$, $ \bm \xi^{(t)} = \mathcal{I}^{F}_{\bm{\theta} + \mathbf{W}^{(t)}}(\bm{x}^{(t)})$ is the output class token of $\mathcal I^{F}$ and $ \mathcal{I}^{F}_{\bm{\theta} + \mathbf{W}^{(t)}} $ represents $ \mathcal{I}^{F}_{\bm \theta} $ augmented with the LoRA block $ \mathbf{W}^{(t)} $ corresponding to $\mathcal T_t$ and $\mathcal Y^{see}$ denotes the set of seen classes. To provide prior experiential references for Anomaly Detection for more accurate anomaly detection, we need to accumulate classification experience $\mathcal{W} $ online within the data stream. To this end, we utilize EMA to accumulate the confidence and the standard deviation of confidence online for input samples, avoiding the impact of noise fluctuations on the estimation error of the mean and standard deviation within a single batch:
\begin{align}
    \mathcal{W}^{e} = &\beta^{e} * \frac{1}{|\mathcal B|}\sum_{i=1}^{|\mathcal B|}c^{(t)}_i + (1-\beta^{e}) * \mathcal{W}^{e}, \\ 
    \mathcal{W}^{std} &= \beta^{std} * \sqrt{\frac{1}{|\mathcal B|}\sum_{i=1}^{|\mathcal B|}\left(c^{(t)}_i-\frac{1}{|\mathcal B|}\sum_{i=1}^{|\mathcal B|}c^{(t)}_i\right)} \\
    &+ (1-\beta^{std})*\mathcal W^{std},
\end{align}
where $\mathcal B = \{(\bm x^{t}, y^{(t)})\}_{i=1}^{|\mathcal B|}$ denotes the input training batch and $\beta^e,\beta^{std}$ are hyperparameters. Thus the online experience can then be represented as $\mathcal W = \{\mathcal{W}^{e}, \mathcal{W}^{std}\}$. To ensure the accuracy of detection during the inference phase, we initialize a learnable temperature scaling parameter $\tau^{y}=1$ for each $y \in \mathcal Y^{see}$. Subsequently, $\tau^{y}$ is dynamically adjusted based on the feedback from the online classification experience. Specifiacally, for each training batch, a pre-detection is conducted before training step using $\mathcal{W}^{e}$ and $\mathcal{W}^{std}$ to filter out high-confidence positive and negative example sets $\mathcal B^{HP}, \mathcal B^{HN}$, as well as low-confidence positive and negative example sets $\mathcal B^{LP}, \mathcal B^{LN}$:
\begin{align}
    &\mathcal B^{HP} = \{\bm z^{(t)}_i\in\mathcal B\mid \tilde{c}^{(t)}_i>\gamma_p, \hat{y}_i^{(t)}=y_i^{(t)} \},\\
    &\mathcal B^{HN} = \{\bm z^{(t)}_i\in\mathcal B\mid \tilde{c}^{(t)}_i>\gamma_p, \hat{y}_i^{(t)}\ne y_i^{(t)} \},\\
    &\mathcal B^{LP} = \{\bm z^{(t)}_i\in\mathcal B\mid \tilde{c}^{(t)}_i<\gamma_n, \hat{y}_i^{(t)}=y_i^{(t)} \},\\
    &\mathcal B^{LN} = \{\bm z^{(t)}_i\in\mathcal B\mid \tilde{c}^{(t)}_i<\gamma_n, \hat{y}_i^{(t)}\ne y_i^{(t)} \},
\end{align}
where $\tilde{c}^{(t)}_i = (c_i^{(t)}-\mathcal W^e)/\mathcal W^{std}$ and $\gamma_p, \gamma_n$ are pre-defined detection threshold. 
Then the dynamic adjustment of $\tau^y$ is executed through feedback from the pre-detection: 
\begin{equation}
    \tau^y = \left(\frac{\kappa}{1 + \exp\left(-\frac{|\mathcal B^{LN}| + |\mathcal B^{HP}|}{|\mathcal B^{HN}| + |\mathcal B^{LP}|} \right)} + \tau^y\right)/2, 
\end{equation}
where $\kappa>1$ is a scaling intensity hyperparameter. This adjustment imposes a higher temperature to penalize predictions of high-confidence negative samples and low-confidence positive samples, thereby enhancing the accuracy of detection.
Based on the instantiated components mentioned above, we introduce the following prototype alignment loss with adaptive temperature scaling:
\begin{equation}
    \small
    \mathcal L_{TPA} (\mathcal B) =  \frac{1}{|\mathcal B|}\sum_{i=1}^{|\mathcal B|} -\log\frac{\left(\left\langle \bm \xi^{(t)}_i , \bm p_w^{y_i^{(t)}}  \right\rangle + \bm p_b^{y_i^{(t)}} \right) \big/ \tau^{y_i^{(t)}}}{\sum_{y'\in\mathcal Y^{\mathcal B}} \left(\left\langle \bm \xi^{(t)}_i, \bm p_w^{y'}  \right\rangle + \bm p_b^{y'} \right) \big/ \tau^{y'}},
\end{equation}
where $ \mathcal Y^{\mathcal B} $ denotes the set of classes contained within batch $ \mathcal B $. After learning on the objective, the action $ \mathcal{A}_L(\mathcal{S}_t^{tr}, \mathcal{I}^{F}, \mathcal{I}^{S}) $ returns the updated $ \mathcal{M}^{STM} $, $ \mathcal{W} $, and $ \mathcal{M}^{Sce} $.

\subsection{Memory Restructuring Based on Scenario Replay}
After learning STM, it is necessary to restructuring STM according to the previous scenario memory, integrating fragmented task memories into a task-general LTM. Since the scenarios in the episodic description pool returned by $ \mathcal{A}_L $ are stored in text form, to replay them as scenario images, we introduce a pre-trained SDXL \cite{abs-2307-01952} to perform the scenario reproduction operation $\tilde{\mathcal{R}}_{\mathcal{M}}$. We first randomly sample $ N $ descriptions $\{\bm{s}_i\}_{i=1}^{N}$ from the description pool $ \mathcal{M}^{Sce} $, then execute the reproduction, and in conjunction with the corresponding class labels of these descriptions to construct a set of scenarios $ \tilde{\mathcal S} = \{\tilde{\bm{x}}_i, y_i\}_{i=1}^{N} = \{\tilde{\mathcal{R}}_{\mathcal{M}}(\bm{s}_i)\}_{i=1}^{N} $.
Subsequently, based on the scenario set, we perform memory restructuring utilizing the task-specific LoRAs. Drawing inspiration from the LoRA-hub's strategy of combining multiple task LoRAs for compositional generalization \cite{abs-2307-13269}, we conduct composite optimization on $\bm \alpha = (\alpha_1, \alpha_2, \dots, \alpha_t)$ over the scenario set targeting the objective $\mathcal L_{TPA}$.
\begin{equation}
\scriptsize
    \bm\alpha_{\Sigma} = \min_{\bm \alpha} \frac{1}{|\tilde{\mathcal S}|}\sum_{i=1}^{|\tilde{\mathcal S}|} -\log\frac{\left(\left\langle \mathcal{I}^{F}_{\bm{\theta} + \widehat{\mathbf{W}}_{\Sigma}}(\bm{x}^{(t)}) , \bm p_w^{y_i}  \right\rangle + \bm p_b^{y_i} \right) \big/ \tau^{y_i}}{\sum_{y'\in\mathcal Y^{\tilde{\mathcal S}}} \left(\left\langle \mathcal{I}^{F}_{\bm{\theta} + \widehat{\mathbf{W}}_{\Sigma}}(\bm{x}^{(t)}), \bm p_w^{y'}  \right\rangle + \bm p_b^{y'} \right) \big/ \tau^{y'}},
\end{equation}
where $\widehat{\mathbf W}_{\Sigma}=(\sum_{i=1}^{t} \alpha_i \mathbf B^{(i)})(\sum_{i=1}^{t}\alpha_i \mathbf A^{(i)})$. 
Upon obtaining $\bm{\alpha}_i^\Sigma$, we proceed to merge memories to derive the reorganized parameter memory $\mathbf{W}_\Sigma = (\sum_{i=1}^{t} \alpha_i^{\Sigma} \mathbf{B}^{(i)})(\sum_{i=1}^{t} \alpha_i^{\Sigma} \mathbf{A}^{(i)})$, where $\alpha_i^\Sigma$ is the $i$-th element of $\bm{\alpha}_{\Sigma}$. Subsequently, we consolidate the restructured parameter memory and representation memory into $\mathcal M^{LTM}$: $\mathcal M^{LTM}_p = \mathbf W_{\Sigma}$, $\mathcal M^{LTM}_r = \mathcal M^{LTM}_r \cup \bigcup_{y \in \mathcal Y^{(t)}} \{\bm p_w^y, \bm p_b^y\}$. When the size of $|\mathcal M^{STM}_p|$ reaches the restructuring period $e$, we perform parameter memory initialization: $\mathcal M^{STM}_p = \mathbf{W}_\Sigma$ to control the maximum number of short-term parameter memories to be at most $e$, thereby ensuring a fixed storage requirement for the parameter memories. After memory restructuring, the action $\mathcal A_{R}(\mathcal M^{Sce}, \mathcal R_{\mathcal M}, \mathcal M^{STM}, \mathcal M^{LTM})$ returns the restructured LTM \(\mathcal M^{LTM}\).

\subsection{Fast and Slow Inference Based On Classification Experience}

During the inference stage, for each test sample $\bm x$, the long-term parameter memory is first integrated into $\mathcal I^{F}$ and fast inference is executed: $\bm \xi = \mathcal I^{F}_{\bm\theta+\mathbf W_{\Sigma}}(\bm x)$, and the predictive confidence is calculated as $c = \max_{1\le i\le |\mathcal Y^{see}|}\exp(\langle\bm \xi, \bm p_w^y\rangle+\bm p_b^y)/\sum_{y'\in\mathcal Y^{see}}\exp(\langle\bm \xi, \bm p_w^{y'}\rangle+\bm p_b^{y'})$. Building upon this, we propose CA-OAD to perform anomaly detection in conjunction with accumulated classification experience. If $\tilde{c}^{(t)} = (c^{(t)} - \mathcal W^e) / \mathcal W^{std} < \gamma_n$, then proceed with the slow inference $\mathcal I^{S}$; otherwise, retain the prediction of $F^{S}$, denote as $\hat y$. Specifically, based on the TopK labels $\hat{y}^{TopK}$ output by the fast inference, we have designed a multiple-choice question as a query $\mathcal Q(\hat{y}^{TopK})$:

\begin{mytexttbox}
\textbf{Question}: Analyze the given picture and select the most likely class it belongs to from the options provided below. Please choose only one class.

\textbf{Choices}: $\text{Textualize}(\hat{y}^{TopK}_1)$, $\text{Textualize}(\hat{y}^{TopK}_2)$, $\dots$, $\text{Textualize}(\hat{y}^{TopK}_K)$.
\end{mytexttbox}
Here, 'Textualize' performs the textual representation of the labels. Subsequently, we have the MLLM executing slow inference to answer the constructed query, yielding $\hat{\hat{y}} = \mathcal I^{S}(\mathcal Q(\hat{y}^{TopK}))$, and we conduct label correspondence operations on the returned results. If the returned result contains only one of the given classes, it is considered a valid response, otherwise, we retain the result of the fast inference, i.e. $\hat{\hat{y}}=\hat{y}$. After the inference is concluded, the action $\mathcal A_{\mathcal I}(\mathcal I^{F}, \mathcal I^{S}, \mathcal I^{Swi}, \mathcal W)$ returns the final prediction result $\hat{\hat{y}} = \hat{y}$.

\subsection{Workflow of SR-CIS} 
Integrating the aforementioned components, we have constructed the SR-CIS, which achieves stable memory preservation and restructuring through the decoupling of memory along with inference. It stores scenario memory through textual descriptions and periodically initializing parameter memory, ensuring limited storage requirements. The incorporation of memory restructuring ensures the classification backbone's flexibility to adapt to tasks, thus providing plasticity. Meanwhile, there is no need for extensive data resources for online generative model training. The workflow of SR-CIS is depicted in Fig. \ref{Workflow}, for each task $\mathcal T_t$, we first execute $\mathcal A_L(\mathcal S_{t}^{(tr)}, \mathcal I^{F}, \mathcal I^{S})$ to learn and store $\mathcal M^{STM}$, while accumulating online experience $\mathcal W$ and $\mathcal M^{Sce}$. Upon completion of training for $\mathcal T_t$, we perform $\mathcal A_{R}(\mathcal M^{Sce}, \mathcal R_{\mathcal M}, \mathcal M^{STM}, \mathcal M^{LTM})$ for memory restructuring, updating the LTM $\mathcal M^{LTM}$. Ultimately, on the evaluation set $\mathcal E_t$, leveraging the online experience $\mathcal W$, we perform efficient switch between fast and slow inferences through CA-OAD to execute \(\mathcal A_{\mathcal I}(\mathcal I^{F}, \mathcal I^{S}, \mathcal I^{Swi}, \mathcal W)\) to get the final prediction result.
\section{Experiments}
% \vspace{-10pt}
\begin{table*}[ht]
  \centering
   \caption{Results (\%) on CIFAR100, DomainNet and ImageNet-R. We report results over 5 trials. The bold font is utilized to denote the optimal results, the underline is applied to indicate the suboptimal outcomes. For SR-CIS the results with (w/) and without (w/o) memory restructuring are documented.
  }
  \setlength{\tabcolsep}{10pt}  
  \resizebox{0.95\textwidth}{!}{
  \begin{tabular}{l c c c c c c} 
  \toprule[1.5pt] 
  \textbf{Tasks} & \multicolumn{2}{c}{\textbf{CIFAR100}} & \multicolumn{2}{c}{\textbf{DomainNet}} & \multicolumn{2}{c}{\textbf{ImageNet-R}}\\
  \midrule
  \rule{0pt}{10pt} \textbf{Method} & $\bm{ACC_{10}}$ ($\uparrow$) & $\bm{a_{10,10}}$ ($\uparrow$) & $\bm{ACC_{5}}$ ($\uparrow$) & $\bm{a_{5,5}}$ ($\uparrow$) & $\bm{ACC_{10}}$ ($\uparrow$) & $\bm{a_{10,10}}$ ($\uparrow$) \\
  \midrule[1.5pt]
  \emph{joint}  
  & $91.92 \pm 0.5$ & - 
  & $77.72 \pm 0.4$ & - 
  & $81.14 \pm 0.34$ & -  \\ 
  \emph{sequential}   
  & $58.74 \pm 3.59$ & $97.80 \pm 2.3$       
  & $30.00 \pm 1.21$ & $68.78 \pm 3.3$ 
  & $46.72 \pm 1.21$ & $91.91 \pm 3.3$\\
  \hline
  ICL~\cite{abs-2403-02628}
  & $79.25 \pm 2.3$ & $77.90 \pm 2.5$
  & $41.18 \pm 1.5$ & 55.32 $\pm$ 1.3 
  & $51.82 \pm 1.8$ & $56.22 \pm 1.4$\\
  L2P~\cite{0002ZL0SRSPDP22}
  & $79.68 \pm 2.6$ & $81.60 \pm 3.1$
  & $49.86 \pm 1.9$ & $49.16 \pm 1.3$ 
  & $48.05 \pm 3.3$ & $44.93 \pm 3.7$\\
  DualPrompt~\cite{0002ZESZLRSPDP22}
  & \underline{80.78 $\pm$ 3.9} & $82.00 \pm 1.7$
  & $53.59 \pm 1.5$ & $51.73 \pm 1.6$ 
  & $55.99 \pm 3.2$ & $54.47 \pm 5.3$\\
  % \hline
  CODA-P~\cite{SmithKGCKAPFK23}
  & $80.32 \pm 2.3$ & \underline{82.60 $\pm$ 2.7}
  & $57.91 \pm 2.4$ & $54.91 \pm 1.7$ 
  & $56.52 \pm 4.6$ & $54.57 \pm 3.7$\\ 
  LAE~\cite{gao2023unified}
  & $80.47 \pm 2.0$ & $82.20 \pm 1.3$
  & $55.59 \pm 4.3$ & $51.93 \pm 2.8$ 
  & $59.45 \pm 4.5$ & $55.95 \pm 2.0$\\
  % \hline
  \mbox{InfLoRA}~\cite{abs-2404-00228}
  & $78.67 \pm 2.4$ & 78.30 $\pm$ 2.9
  & \underline{58.59 $\pm$ 2.3} & \underline{56.64} $\pm$ 1.9
  & 46.65 $\pm$ 1.5 & 45.69 $\pm$ 2.8\\ 
  \rowcolor{gray!20} \mbox{SR-CIS (w/o restructuring)} 
  & 80.42 $\pm$ 2.1  & 82.24 $\pm$ 1.5 
  & 57.78 $\pm$ 2.5 & 55.68 $\pm$ 3.0
  & \underline{64.01 $\pm$ 2.3} &  \underline{62.13 $\pm$ 2.7}
  \\
  \rowcolor{gray!20} \mbox{SR-CIS (w/ restructuring)} 
  & \textbf{85.85} $\pm$ \textbf{2.4}  & \textbf{83.90} $\pm$ \textbf{2.6} 
  & \textbf{60.99} $\pm$ \textbf{1.6} & \textbf{57.68} $\pm$ \textbf{1.8}
  & \textbf{71.27} $\pm$ \textbf{2.1} &  \textbf{77.55} $\pm$ \textbf{1.3}
  \\
  \bottomrule[1.5pt]
  \end{tabular}}
  \label{tab:common}
  \vskip -0.1in
\end{table*}

\begin{table*}[t]
  \centering
  \setlength{\tabcolsep}{10pt}  
  \caption{Results (\%) on CIFAR100, DomainNet and ImageNet-R on few-Shot setting. We report results over 5 trials. The bold font is utilized to denote the optimal results, the underline is applied to indicate the suboptimal outcomes. For SR-CIS the results with (w/) and without (w/o) memory restructuring are documented.%
  }
  \resizebox{0.95\textwidth}{!}{
  \begin{tabular}{l c c c c c c} 
  \toprule[1.5pt] 
  \textbf{Tasks} & \multicolumn{2}{c}{\textbf{CIFAR100}} & \multicolumn{2}{c}{\textbf{DomainNet}} & \multicolumn{2}{c}{\textbf{ImageNet-R}}  \\
  \hline
  \rule{0pt}{10pt} \textbf{Method} & $\bm{ACC_{10}}$ ($\uparrow$) & $\bm{a_{10,10}}$ ($\uparrow$) & $\bm{ACC_{5}}$ ($\uparrow$) & $\bm{a_{5,5}}$ ($\uparrow$) & $\bm{ACC_{10}}$ ($\uparrow$) & $\bm{a_{10,10}}$ ($\uparrow$) \\
  \midrule[1.5pt]
  \emph{joint}  
  & $53.41 \pm 0.85$ & - 
  & $61.86 \pm 0.74$ & - 
  & $53.13 \pm 1.04$ & - \\ 
  \emph{sequential}   
  & $20.23 \pm 3.59$ & $14.50 \pm 1.23$       
  & $14.12 \pm 1.21$ & $35.16 \pm 4.7$ 
  & $18.71 \pm 1.21$ & $27.76 \pm 5.5$ \\
  \hline
  ICL~\cite{abs-2403-02628}
  & $33.27 \pm 2.3$ & $ 23.90 \pm 2.5$
  & $19.92 \pm 1.5$ & \underline{$ 34.33 \pm 2.3$} 
  & $25.85 \pm 1.8$ & $ 21.10 \pm 1.4$ \\
  L2P~\cite{0002ZL0SRSPDP22}
  & $31.37 \pm 2.6$ & $ 36.30 \pm 3.1$
  & $21.85 \pm 1.9$ & $ 26.25 \pm 1.3$
  & $19.56 \pm 3.3$ & $ 16.98 \pm 3.7$ \\
  DualPrompt~\cite{0002ZESZLRSPDP22}
  & $35.57 \pm 3.9$ & $30.50 \pm 1.7$
  & $22.88 \pm 1.5$ & $29.99 \pm 1.6$
  & $21.43 \pm 3.2$ & $18.11 \pm 5.3$ \\
  % \hline
  CODA-P~\cite{SmithKGCKAPFK23}
  & $27.58 \pm 2.3$ & $37.20 \pm 3.7$
  & $23.90 \pm 2.4$ & $25.06 \pm 1.7$
  & \underline{$26.73 \pm 4.6$} & $20.92 \pm 3.7$ \\ 
  LAE~\cite{gao2023unified}
  & \underline{$46.06 \pm 2.0$} & \underline{$ 43.90 \pm 1.3$}
  & $21.18 \pm 4.3$ & $30.43 \pm 2.8$
  & $26.50 \pm 4.5$ & \underline{$21.20 \pm 2.0$} \\
  \mbox{InfLoRA}~\cite{abs-2404-00228}
  & $35.25 \pm 2.1$ & 31.40 $\pm$ 2.8
  & 19.77 $\pm$ 2.3 & 29.57 $\pm$ 3.7
  & 19.23 $\pm$ 2.3 & 18.75 $\pm$ 2.6 \\ 
  \rowcolor{gray!20} \mbox{SR-CIS (w/o restructuring)} 
  & 45.15 $\pm$ 2.1  & 36.57 $\pm$ 1.5 
  & \underline{31.04 $\pm$ 2.5} & 30.26 $\pm$ 3.0
  & 26.28 $\pm$ 2.3 &  21.09 $\pm$ 2.7\\
  \rowcolor{gray!20} \mbox{SR-CIS (w/ restructuring)} 
  & \textbf{52.44} $\pm$ \textbf{3.2}  & \textbf{45.68} $\pm$ \textbf{2.1} 
  & \textbf{34.04} $\pm$ \textbf{2.3} & \textbf{35.32} $\pm$ \textbf{2.6}
  & \textbf{35.28} $\pm$ \textbf{1.5} &  \textbf{26.09} $\pm$ \textbf{1.8}\\
  \bottomrule[1.5pt]
  \end{tabular}}
  \label{tab:Fewshot}
  \vskip -0.1in
\end{table*}

\begin{figure*}[ht]
    \centering
    \includegraphics[width=0.9\textwidth]{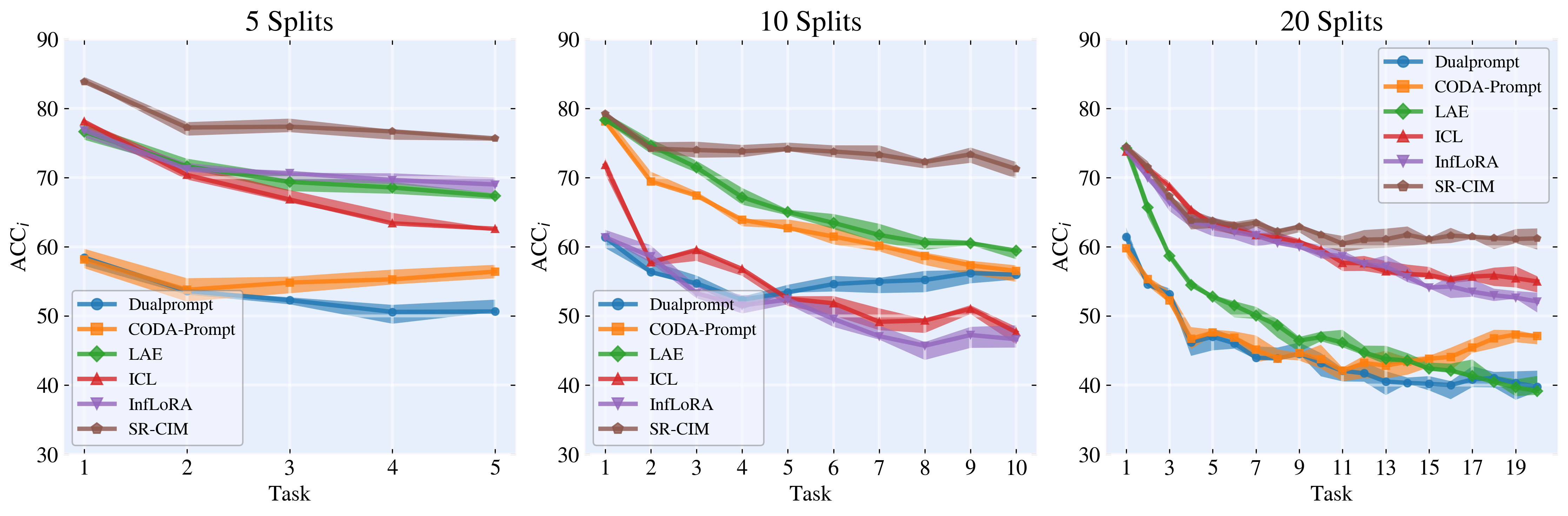}
    \vspace{-10pt}
    \caption{The results on distinct task splits.}
    \label{Fig:Tasks}
    \vspace{-10pt}
\end{figure*}

\subsection{Experimental Settings}
Aligned with current baselines based on fine-tuning pre-trained ViTs \cite{SmithKGCKAPFK23, abs-2404-00228}, we employ CIFAR100 \cite{Krizhevsky2009LearningML}, ImageNet-R \cite{HendrycksBMKWDD21}, and DomainNet \cite{PengBXHSW19} for model training and evaluation. CIFAR100 comprises 100 classes, each with 500 training images and 100 testing images, serving as a common dataset in continual learning scenarios. ImageNet-R is an extension of the ImageNet dataset, encompasses 200 classes totaling 30,000 images, with 20\% allocated for testing. It has emerged as a benchmark dataset for continual learning tasks \cite{gao2023unified, 0002ZESZLRSPDP22}. DomainNet, a dataset of common objects across 6 diverse domains, includes 345 classes with 600,000 images. We constitute our training and testing sets by taking the first 200 images from each class in the training set and the first 100 images from each class in the test set.

We follow the setting of \cite{SmithKGCKAPFK23, abs-2404-00228} and divide CIFAR100 into 10 tasks, each containing 10 classes. ImageNet-R is divided into 5, 10, and 20 tasks, each containing 40, 20, and 10 classes, respectively. DomainNet is divided into 5 tasks, each containing 69 classes.

We use existing continuous learning methods \cite{0002ZL0SRSPDP22, abs-2404-00228} to evaluate model performance using two popular metrics: the final average accuracy ($ACC_T$) and the final current task accuracy ($a_{T,T}$). Here, $ACC_i$ is defined as $ACC_i=\frac{1}{i}\sum_{j=1}^i a_{i,j}$, where $a_{i,j}$ represents the accuracy of the $j$-th task once the model has learned the $i$-th task.

\noindent\textbf{Baselines}
We compare our method with state-of-the-art continuous learning methods, including ICL \cite{abs-2403-02628}, L2P \cite{0002ZL0SRSPDP22}, DualPrompt \cite{0002ZESZLRSPDP22}, CODA-P \cite{SmithKGCKAPFK23}, LAE \cite{gao2023unified}, and InFLoRA \cite{abs-2404-00228}. Aligning with \cite{abs-2404-00228,0002ZESZLRSPDP22}, we include two naive methods: joint and sequential. The joint method, which involves learning all tasks simultaneously, is generally regarded as the upper bound of continuous learning performance. The sequential method, which uses a backbone network to learn tasks continuously, is considered the lower bound of performance.

\noindent\textbf{Architecture and Training Details} 
We follow the existing work settings \cite{abs-2404-00228} for our experiment. Specifically, we use the ViT-B/16 model pre-trained on ImageNet 1K. All methods adhere to the original paper's settings. We adopt an online continuous learning setting, where each task trains for only one epoch. The batch size is set to 10 for all experiments. We set the buffer size to 600 for ICL and 0 for all other methods. For more experimental details, please refer to the Appendix \ref{details}.

\subsection{Experimental Results}
\textbf{Comparison on Standard Sequential Tasks.} 
Table \ref{tab:common} presents the results of comparative experiments with SR-CIS configured as follows: restructuring period $e=3$, temperature scaling hyperparameter $\kappa = 1.2$, CA-OAD threshold $\gamma = 1.28$, and EMA parameters $\beta^e$ and $\beta^{std}$ both set to $0.01$. Outcomes are documented for SR-CIS with (w/) and without (w/o) memory restructuring (incremental addition of LoRA for new tasks). SR-CIS shows remarkable performance, surpassing other baselines. On ImageNet-R, it exceeds the second-best baseline, LAE, by 11.82\%. Additionally, it achieves a single-task accuracy on the final task that is 21.33\% higher than the second best method, ICL. Furthermore, SR-CIS with memory restructuring significantly outperforms its counterpart without memory restructuring in both final average accuracy and final current task accuracy, highlighting the efficacy of memory restructuring in enhancing the model's memory stability and plasticity for new tasks.

\noindent\textbf{Comparison on Few-Shot Sequential Tasks.}
To validate our method's performance in low-data resources scenarios, we extend to a Few-Shot setting and conduct comparative experiments. Specifically, we limit the training data to the first 20 images per class from each of the three datasets, adopting a 20-shot scenario while keeping all other settings unchanged. The experimental results in Table \ref{tab:Fewshot} demonstrate that SR-CIS maintains a clear advantage in both average precision and current-task precision. This indicates that SR-CIS retains good plasticity and memory stability even when data resources are scarce.

\noindent\textbf{Validation across varying task lengths}. To examine the memory stability of SR-CIS under different task lengths, following \cite{SmithKGCKAPFK23,abs-2404-00228}, we configure three distinct task splits on ImageNet-R consisting of 5, 10, and 20 tasks respectively, and conduct comparative experiments. We record the average precision at the conclusion of each task epoch and plot forgetting curves, which are depicted in Figure \ref{Fig:Tasks}. SR-CIS significantly exhibits the highest final average accuracy across all three task splits, demonstrating a lower forgetting rate. Even under the 20 Splits setup, SR-CIS still achieves an over 10\% boost in final average accuracy compared to the second-best method, ICL, firmly validating SR-CIS's superior memory stability.

\begin{figure}
    \centering
    \includegraphics[width=0.75\linewidth]{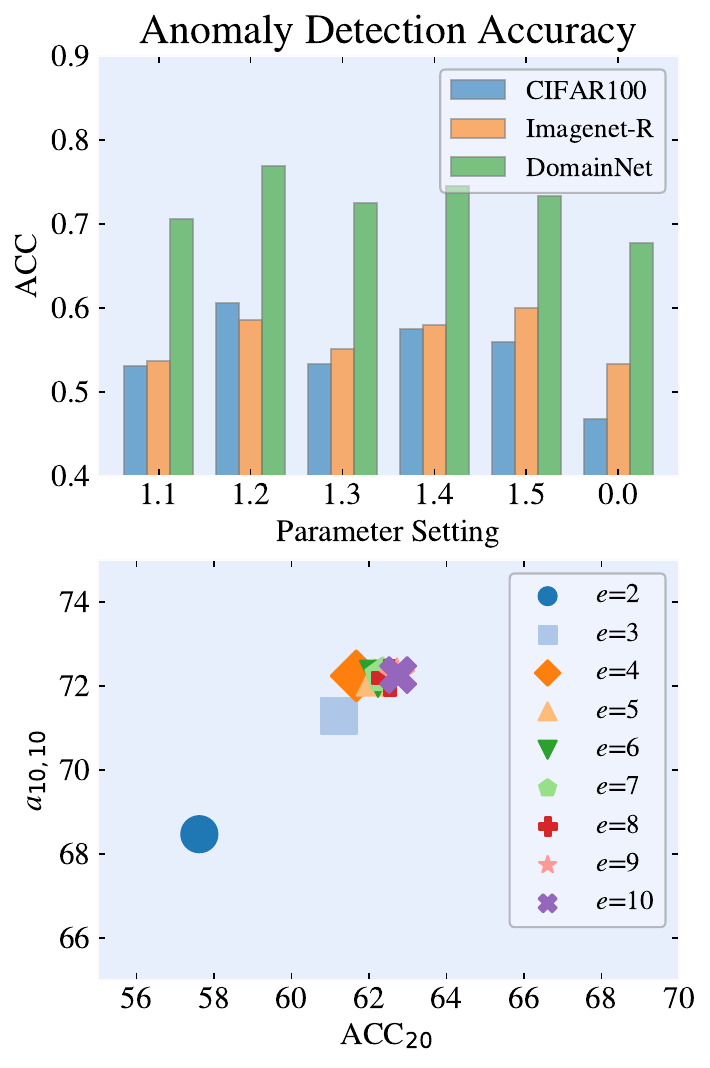}
    \vspace{-10pt}
    \caption{Upper: The detection accuracy on different $\kappa$. "0" denote w/o adaptive temperature scaling. Lower: SR-CIS performence on 20-split ImageNet-R with different choice of $e$.}
    \label{fig:abl}
    \vspace{-10pt}
\end{figure}

\noindent\textbf{The Impact of $\kappa$ on Anomaly Detection Accuracy}. 
To investigate the effect of the temperature scaling factor $\kappa$ on the anomaly detection accuracy of CA-OAD, we record the proportion of samples misclassified by $\mathcal I^F$ across three datasets under various $\kappa$ configurations. We also document the results without adaptive temperature scaling during training. As shown in Figure \ref{fig:abl}, the detection accuracy with adaptive temperature scaling significantly surpasses that without it. Specifically, CA-OAD achieves a remarkable improvement in detection accuracy when $\kappa=1.2$, demonstrating the efficacy of adaptive temperature adjustment in enhancing anomaly detection capabilities.

\noindent\textbf{The Impact of Memory Restructuring Period on CIL Performance of SR-CIS}. 
The memory restructuring epoch $e$ dictates the quantity of parameter memories retained. We employ varied restructuring periods to explore the trade-off between SR-CIS's CIL performance and the restructuring period on extended task sequences, using a 20 Splits task configuration on ImageNet-R. The results, shown in Figure \ref{fig:abl}, reveal a marked rise in both the average accuracy and final current task accuracy of SR-CIS when the period $e$ increases from 2 to 3. However, beyond a restructuring cycle of $e=3$, SR-CIS's incremental performance exhibits minimal further improvement, suggesting that $e=3$ suffices for effective memory restructuring. This ensures that, under the constraint of limited parameter memory, SR-CIS can continue to adapt effectively to prolonged sequences of tasks.
\section{Conclusion}

In this paper, inspired by CLS theory, we propose the SR-CIS framework with decoupled memory and reasoning to enable self-reflective memory evolution and continual learning. SR-CIS deconstructs and collaborates short-term and long-term memory, as well as fast and slow inference, by constructing decomposed CIM and CMM. Our designed CA-OAD mechanism ensures accurate detection of hard samples and efficient switching between fast and slow inference. Meanwhile, CMM deconstructs parameter memory and representation memory through LoRA and prototype weights and biases, executing parameter memory combination and periodic LTM restructuring via a scenario description pool. Balancing model plasticity and memory stability under limited storage and data resources, SR-CIS surpasses current competitive baselines in multiple standard and low-sample incremental learning experiments, providing a systematic perspective for future research on better understanding and modeling human learning mechanisms.

{
    \small
    \bibliographystyle{ieeenat_fullname}
    \bibliography{main}
}
\clearpage
\setcounter{page}{1}
\maketitlesupplementary

\section{Case of the scenario description and replay}
\begin{figure}[h]
    \centering
    \includegraphics[width=0.48\textwidth]{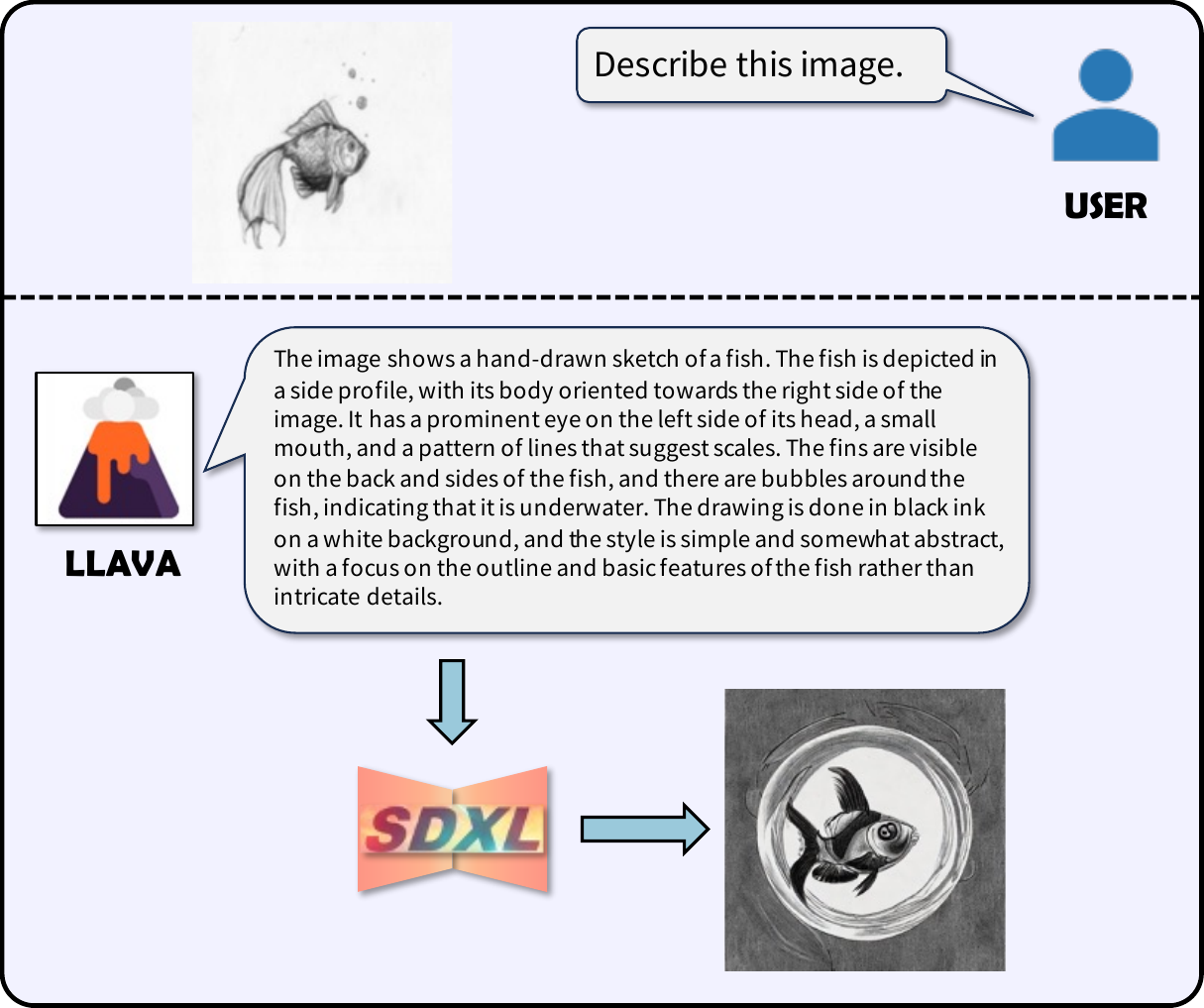}
    \caption{LLaVA generates a description of this image and puts it into the stable diffusion demo to generate the image.}
    \label{fig:case-llava}
\end{figure}

Figure \ref{fig:case-llava} provides an example to understand how our scenario description and scenario replay work: by asking LLAVA to describe the input image to generate a scenario description, and then in the subsequent scenario replay, SDXL comprehends the description and generates a replay image for memory restructuring.

\section{More Experimental Details}
\label{details}
For all method, we resize all images to $256 \times 256$, then use CenterCrop for all images to $224 \times 224$. We take 1.28, the 90\% quantile of the normal distribution as the threshold, which means we retrieve difficult to classify samples with a 90\% confidence level. As done in DualPrompt~\cite{0002ZESZLRSPDP22}, we utilized 20\% of the training data as validation data to define the hyperparameters $\beta = \{\beta^e, \beta^{std} \}$ and the lora rank $r$, ultimately determining $\beta^e = \beta^{std} = 1 \times 10^{-3}$ and $r = 32$. We searched for $\beta$ in value of $\{1e^{-3},1e^{-2},1e^{-1},1\}$.For other methods, we have provided the hyperparameters and their settings mentioned in the original paper. We set the learning rate to 1e-3, and the detailed parameters are shown in the Table \ref{tab:hyperparameters}. All models are implemented using PyTorch on a single NVIDIA A800 GPU.

\begin{table}[t!]
  \centering
  \scriptsize
  \centering
  \caption{List of hyper-parameters for different methods. The meaning of different hyperparameters is given in Section. The hyperparameter $\epsilon$ in InfLoRA is explained in Section}
  \vspace{\baselineskip} 
  \begin{tabular}{ll}
    \toprule[1.5pt]
    \textbf{Methods}     & \textbf{Hyper-Parameters} \\
    \midrule
    L2P       & lr: 0.0001~(ImageNet-R, DomainNet, CIFAR100) \\
              & $l$: 1~(ImageNet-R, DomainNet, CIFAR100) \\
              & $p$: 30~(ImageNet-R, DomainNet, CIFAR100) \\
              & $e$: 20~(ImageNet-R, DomainNet, CIFAR100) \\
    \midrule
    DualPrompt & lr: 0.0001~(ImageNet-R, DomainNet, CIFAR100) \\
               & $l_{E}$: 3~(ImageNet-R, DomainNet, CIFAR100) \\
               & $l_{S}$: 2~(ImageNet-R, DomainNet, CIFAR100) \\
               & $e_{E}$: 20~(ImageNet-R, DomainNet, CIFAR100) \\
               & $e_{S}$: 6~(ImageNet-R, DomainNet, CIFAR100) \\
    \midrule
    CODA-P       & lr: 0.0001~(ImageNet-R, DomainNet, CIFAR100) \\
                 & $l$: 5~(ImageNet-R, DomainNet, CIFAR100) \\
                 & $p$: 100~(ImageNet-R, DomainNet, CIFAR100) \\
                 & $e$: 8~(ImageNet-R, DomainNet, CIFAR100) \\
    \midrule
    LAE       & lr: 0.0001~(ImageNet-R, DomainNet, CIFAR100) \\
              & $r$: 5~(ImageNet-R, DomainNet, CIFAR100) \\
    \midrule
    C-LoRA    & lr: 0.0001~(ImageNet-R, DomainNet, CIFAR100) \\
              & $r$: 64~(ImageNet-R, DomainNet, CIFAR100) \\
              & $\lambda$: 0.5~(ImageNet-R, DomainNet, CIFAR100) \\
    \midrule
    InfLoRA    & lr: 0.0001~(ImageNet-R, DomainNet, CIFAR100) \\
               & $r$: 10~(ImageNet-R, DomainNet, CIFAR100)\\
               & $\epsilon$: $0.98$~(ImageNet-R), $0.95$~(CIFAR100, DomainNet) \\
    \bottomrule[1.5pt]
  \end{tabular}
  \label{tab:hyperparameters}
\end{table}

\end{document}